\documentclass[preprint,10pt]{elsarticle}
\usepackage{setspace}
\usepackage{geometry}
\usepackage{graphicx}
\usepackage{physics}
\usepackage{amsmath}
\usepackage{tikz}
\usepackage{mathdots}
\usepackage{yhmath}
\usepackage{cancel}
\usepackage{color}
\usepackage{siunitx}
\usepackage{array}
\usepackage{multirow}
\usepackage{amssymb}
\usepackage{gensymb}
\usepackage{tabularx}
\usepackage{extarrows}
\usepackage{booktabs}
\usetikzlibrary{fadings}
\usetikzlibrary{patterns}
\usetikzlibrary{shadows.blur}
\usetikzlibrary{shapes}
\usepackage{diagbox}
\usepackage{bbding}
\usepackage{amsthm}
\usepackage{float}
\usepackage{bm}
\usepackage{pdflscape}
\usepackage{rotating}
\usepackage{booktabs}

\theoremstyle{definition}

\journal{Information Fusion}

\begin{document}

\begin{frontmatter}

\title{Graph-Based Adversarial Domain Generalization with Anatomical Correlation Knowledge for Cross-User Human Activity Recognition}

\author[inst1]{Xiaozhou Ye\corref{cor1}\fnref{fn1}}
\ead{xye685@aucklanduni.ac.nz}

\author[inst1]{Kevin I-Kai Wang\corref{cor2}}
\ead{kevin.wang@auckland.ac.nz}
\cortext[cor2]{Corresponding author}

\affiliation[inst1]{organization={Department of Electrical, Computer, and Software Engineering, The University of Auckland}, city={Auckland}, country={New Zealand}}

\begin{abstract}
Cross-user variability poses a significant challenge in sensor-based Human Activity Recognition (HAR) systems, as traditional models struggle to generalize across users due to differences in behavior, sensor placement, and data distribution. To address this, we propose \textbf{GNN-ADG} (Graph Neural Network with Adversarial Domain Generalization), a novel method that leverages both the strength from both the Graph Neural Networks (GNNs) and adversarial learning to achieve robust cross-user generalization. GNN-ADG models spatial relationships between sensors on different anatomical body parts, extracting three types of \textbf{Anatomical Units}: (1) \textbf{Interconnected Units}, capturing inter-relations between neighboring sensors; (2) \textbf{Analogous Units}, grouping sensors on symmetrical or functionally similar body parts; and (3) \textbf{Lateral Units}, connecting sensors based on their position to capture region-specific coordination. These units information are fused into an unified graph structure with a cyclic training strategy, dynamically integrating spatial, functional, and lateral correlations to facilitate a holistic, user-invariant representation. Information fusion mechanism of GNN-ADG occurs by iteratively cycling through edge topologies during training, allowing the model to refine its understanding of inter-sensor relationships across diverse perspectives. By representing the spatial configuration of sensors as an unified graph and incorporating adversarial learning, Information Fusion GNN-ADG effectively learns features that generalize well to unseen users without requiring target user data during training, making it practical for real-world applications. Evaluations on OPPORTUNITY and DSADS datasets show that Information Fusion GNN-ADG outperforms state-of-the-art methods in cross-user activity recognition. By fusing biomechanical insights with machine learning techniques, our approach bridges the gap between domain knowledge and computational models, offering a robust solution to the challenge of cross-user variability in HAR.
\end{abstract}


\end{frontmatter}

\section{Introduction}

Human Activity Recognition (HAR) has become increasingly important due to its wide range of applications in healthcare, sports analytics, smart homes, and human-computer interaction \cite{bulling2014tutorial}. Wearable sensors provide rich and continuous data streams that can be used to recognize and classify human activities \cite{kwapisz2011activity}. However, one of the primary challenges in HAR is the variability across different users, known as cross-user variability \cite{s24247975}. Differences in body dynamics, individual behaviors, and sensor placements can lead to performance degradation when a model trained on one set of users is applied to another \cite{ye2023cross}. Therefore, developing models that can generalize well across users is crucial for the widespread adoption of HAR.

Traditional HAR approaches typically involve feature extraction followed by classification using models such as Convolutional Neural Networks (CNNs) or Recurrent Neural Networks (RNNs) \cite{ordonez2016deep}. While these models have proven effective, they inherently learn user-specific patterns, leading to decreased performance when applied to unseen users \cite{hammerla2016deep}. Recent research has explored domain adaptation and transfer learning techniques \cite{wang2018deep} to address this challenge. However, these methods often require labeled data from the target user, limiting their practicality in real-world scenarios where acquiring such data is impractical.

Domain generalization has been proposed to handle scenarios where no data from the target user is available during training \cite{li2018learning}. While promising, many current domain generalization methods struggle to capture the shared relationships between sensor positions across different users \cite{ahmad2021graph}. This limitation arises because they focus primarily on individual user characteristics, overlooking the common patterns inherent in human anatomy and movement coordination. Even though individuals vary in terms of gender, weight, height, and behavioral tendencies, certain anatomical correlations between body parts remain consistent across users.

For example, during activities like running, specific anatomical correlations are universally present: the left and right shoulders move in opposite directions to balance arm movements, and the legs alternate in motion to enable locomotion. Capturing these intrinsic patterns of coordinated movement across different anatomical parts is essential for improving the generalization capability of HAR models \cite{ahmad2021graph}.

To address these challenges, we propose a \textbf{Sensor Position-Aware Graph Neural Network with Adversarial Domain Generalization (GNN-ADG)}. Our approach leverages Graph Neural Networks (GNNs) to explicitly model the spatial relationships between sensors, corresponding to different anatomical parts of the human body, within a unified graph structure. Specifically, we extract three types of anatomical units:

\begin{itemize} \item \textbf{Interconnected Anatomical Units}: Modeling immediate spatial dependencies by connecting sensors on neighboring anatomical parts. \item \textbf{Analogous Anatomical Units}: Grouping sensors on functionally similar or symmetrical body parts to capture bilateral movement patterns. \item \textbf{Lateral Anatomical Units}: Connecting sensors based on their position on the left, right, or middle part of the body to capture region-specific coordination. \end{itemize}

A key innovation of our method is the unified graph with a cyclic training strategy, which integrates these three anatomical perspectives into a single, dynamic framework. Rather than training separate models for each graph type, we cycle through the edge topologies corresponding to each anatomical unit during training. This information fusion process allows the model to iteratively refine its understanding by alternating focus among spatial, functional, and lateral correlations. By dynamically integrating these perspectives, the unified graph ensures that the model captures a holistic, user-invariant representation of human movement, preventing over-specialization to any single anatomical perspective and enhancing generalization across diverse activities and users.

By integrating these anatomical correlations into a graph-based network, our method captures complex inter-sensor relationships due to individual differences. This design allows Information Fusion GNN-ADG to effectively learn user-invariant features that generalize well to unseen users. Furthermore, we integrate adversarial learning using a Gradient Reversal Layer (GRL) to extract user-invariant features, minimizing the influence of user-specific variations and improving cross-user generalization \cite{ganin2015unsupervised}. Importantly, our method does not require any data from target users during training, making it practical for real-world applications where collecting new user data is impractical.

The main contributions of this paper are as follows:

\begin{itemize} 
\item \textbf{Fusing Anatomical Correlation Knowledge with Adversarial Learning for Domain Generalization}: We propose a framework that systematically integrates biomechanical insights on anatomical part correlations with adversarial domain generalization techniques, enhancing model robustness in cross-user HAR by aligning domain-invariant feature learning with physiological principles. 
\item \textbf{Novel Graph-based Neural Network Architecture with Unified Fusion}: We propose a new graph-based model structure that unifies anatomical correlations through a cyclic training strategy, dynamically integrating spatial, functional, and lateral relationships to capture complex and common inter-sensor patterns.
\item \textbf{Adversarial Domain Generalization}: Our framework employs adversarial learning to extract user-invariant features, enhancing generalization across users without requiring target user data during training. \end{itemize}

We evaluate our method on two benchmark datasets: \textit{OPPORTUNITY} and \textit{DSADS}. Experimental results demonstrate that Information Fusion GNN-ADG outperforms state-of-the-art methods in cross-user activity recognition tasks, highlighting its effectiveness in generalizing to unseen users. Our approach provides a robust solution to the cross-user variability problem in HAR systems. To the best of our knowledge, this is the first time the shared and common knowledge of anatomical correlation between different body positions has been explored for domain generalization in HAR through the lens of information fusion. 

The rest of the paper is organized as follows. Section~\ref{sec:background} provides background on cross-user variability, graph neural networks, and adversarial domain generalization. Section~\ref{sec:method} details our proposed Information Fusion GNN-ADG method. Section~\ref{sec:experimental_setup} describes the experimental setup and discusses the evaluation results, and Section~\ref{sec:conclusion} concludes the paper and outlines future work.

\section{Background}
\label{sec:background}

\subsection{Cross-User Variability in HAR}

Differences among users in HAR emerge due to variations in sensor positioning \cite{gil2023reducing}, physical movements, and personal behavior traits \cite{soleimani2021cross}. These variations create data distribution shifts that complicate the performance of standard machine learning models \cite{wang2024optimization}, especially in vital fields like healthcare, security monitoring, and intelligent environments, where consistent results across diverse individuals are crucial \cite{gupta2022human}. To overcome this, HAR studies have introduced multiple strategies, primarily divided into fine-tuning \cite{guo2021efficacy}, domain adaptation \cite{ye2025cross}, and domain generalization \cite{bento2023exploring}.

Fine-tuning refers to tweaking a pre-existing model to suit a specific dataset, boosting its accuracy for particular users or settings \cite{thukral2025cross}. Ye et al. \cite{ye2024deep} point out that data distribution changes in HAR hinder conventional models, underlining the value of flexible approaches. Manuel et al. \cite{gil2023reducing} report reduced performance when models encounter unseen users, highlighting fine-tuning’s potential. Recent investigations expand on this technique. For example, Gopalakrishnan et al. \cite{gopalakrishnan2024comparative} assess fine-tuning deep networks for action detection, achieving higher precision in security contexts. Genc et al. \cite{genc2024human} apply fine-tuned CNN-LSTM models to improve activity detection using sensor inputs. Meanwhile, Leite et al. \cite{leite2022resource} examine blending continual learning with fine-tuning to manage evolving activity classification data. These developments showcase fine-tuning’s strengths but also reveal hurdles in scalability and privacy due to its dependence on user-specific data, compounded by the challenge of obtaining labeled data from target users, which is often scarce in real-world cases \cite{YeSensor2024}.

Domain adaptation relies solely on unlabeled data from the target user \cite{farahani2021brief}. It refines models trained on a source domain to excel in a target domain, mitigating cross-user differences by synchronizing the data distributions of both domains \cite{xiaozhou2024genDATR}. Avijoy et al. \cite{chakma2021activity} use adversarial learning to pick the most significant features from various source domains, mapping them to the target domain via perplexity scores that gauge each source’s relevance, allowing the model to emphasize key features. Hu et al. \cite{hu2023swl} propose a sample-weighting method, assigning weights to samples based on their classification and domain discrimination losses, computed through a parameterized network optimized fully via a meta-optimization approach. This approach uses meta-classification loss on pseudo-labeled target samples to craft a custom weighting function for cross-user HAR tasks. Additionally, Napoli et al. \cite{napoli2024benchmark} offer a standard for testing domain adaptation and generalization in smartphone-driven HAR. These efforts demonstrate domain adaptation’s promise but note its limitation of needing target data.

Domain generalization strives to build models that perform effectively on unfamiliar domains without target data, tackling a key shortfall in real-world HAR use \cite{hong2024crosshar}. Bento et al. \cite{bento2023exploring}, for instance, study regularization techniques like Mixup and Sharpness-Aware Minimization for domain generalization in accelerometer-driven HAR. Data augmentation plays a vital role here \cite{li2021simple}, enhancing training data diversity to make models more adaptable to varying domains. Volpi et al. \cite{volpi2018generalizing} illustrate how adversarial data augmentation helps models adjust to new domains by mastering diverse user-specific differences.

A core tactic in domain generalization is invariant feature learning, which targets traits that stay steady across domains \cite{lu2022domaininvariant}. Focusing on these stable traits enables models to adapt to new users without overfitting to specific quirks. Advanced methods have emerged for this purpose. ANDMask \cite{parascandolo2021learning} promotes updates to invariant features across domains while blocking retention of domain-unique traits. AdaRNN \cite{du2021adarnn} handles temporal shifts in time series data, using adaptive techniques to align domain distributions and ensure solid generalization despite varying temporal user patterns. DIFEX \cite{lu2022domaininvariant} merges internal and mutual invariance to enrich feature variety, addressing HAR’s complex cross-domain demands by leveraging intra- and inter-domain consistency for robust feature learning across diverse activity patterns. RSC \cite{huang2020self} removes dominant features during training, pushing reliance on label-linked features likely to remain consistent across domains.

Still, current domain generalization methods struggle to fully grasp all critical invariances across varied users and depend heavily on source domain diversity, which may not mirror unseen target domain variability. This highlights the pressing need for new techniques to detect shared, essential activity patterns—like uniform body part coordination during tasks—focusing on traits universal to individuals for strong generalization, even with limited source domain variety. Therefore, our work focuses on domain generalization by fusing biomechanical knowledge into time series analysis for cross-user HAR. 

\subsection{Graph Neural Networks in HAR}

Graph Neural Networks (GNNs) have emerged as a powerful approach for modeling relational and structured data, distinguishing themselves from traditional neural networks by their ability to capture complex relationships between entities \cite{wu2020comprehensive}. In a GNN, entities are represented as nodes, and their connections as edges, forming a graph that encodes both spatial and functional dependencies \cite{seo2018structured}. This framework is particularly well-suited to Human Activity Recognition (HAR), where activities involve consistent coordination of body parts across individuals—a critical invariant that domain generalization techniques must exploit. In HAR, body parts can be modeled as nodes, with edges representing spatial relationships (e.g., proximity of limbs) or functional correlations (e.g., coordinated movements), offering a natural way to extract shared activity patterns.

GNNs have shown effectiveness in HAR, particularly in skeleton-based tasks where inter-part relationships are essential for recognizing actions. For example, Si et al. \cite{si2019attention} proposed an attention-enhanced GNN for skeleton-based action recognition, emphasizing the simultaneous modeling of spatial and temporal dynamics. Building on this, Shi et al. \cite{shi2019two} introduced two-stream adaptive GNNs to improve HAR by capturing dependencies between body parts. More recently, Wieland et al. \cite{wieland2023tinygraphhar} developed TinyGraphHAR, demonstrating GNNs' capability to distinguish complex activities. These studies highlight GNNs' strength in modeling inter-part coordination. However, their application to domain generalization—especially in sensor-based, cross-user scenarios—remains underdeveloped. This gap is significant, as the consistent coordination of body parts across users could be a key lever for enhancing generalization, particularly when source user diversity is limited. Leveraging this consistency could provide a foundation for creating robust, domain-invariant representations.

Despite their strengths, GNNs alone fall short in addressing domain generalization challenges in HAR, such as variability across users and sensors. To bridge this gap, adversarial learning offers a promising solution by enabling the learning of domain-invariant features. This technique pits a feature extractor against a domain discriminator in a minimax game \cite{ganin2016domain}. The feature extractor seeks to generate representations that confuse the discriminator, while the discriminator aims to identify the domain (e.g., user or sensor) of the input data. A key component in this process is the Gradient Reversal Layer (GRL), introduced by Ganin and Lempitsky \cite{ganin2015unsupervised}. During backpropagation, the GRL reverses the gradients from the domain discriminator, encouraging the feature extractor to eliminate domain-specific information and produce more generalized representations.

Adversarial learning has proven effective in domain generalization across various domains. For instance, Li et al. \cite{li2018adaptive} applied adversarial networks to improve generalization in object recognition, while Yao et al. \cite{yao2017deepsense} used adversarial training in HAR to develop representations invariant to user and sensor differences. By integrating adversarial learning with GNNs, we can combine the structural insights of GNNs—such as inter-sensor or inter-part relationships—with the domain-invariance promoted by adversarial training. This hybrid approach is poised to enhance a model's ability to generalize across diverse users without requiring access to their data during training, tackling both cross-user variability and scalability challenges in HAR.

In summary, GNNs excel at modeling spatial and functional dependencies in HAR, yet their potential for domain generalization, particularly in sensor-based, cross-user contexts, remains largely untapped. This paper aims to address this limitation by leveraging GNNs to uncover essential, shared activity patterns through inter-sensor relationships. By incorporating adversarial learning, we further enhance the model’s generalization capabilities, paving the way for robust, scalable HAR systems that overcome the shortcomings of current domain generalization methods.

\section{Method}
\label{sec:method}

\subsection{Problem Overview}  
\label{sec:problem}  

In the domain generalization setting for cross-user HAR, we consider \( K \) source users with labeled datasets. For each source user \( k \), the dataset is defined as:  
\[  
\mathcal{D}_k^{\text{Source}} = \left\{ \left( \mathbf{x}_i^k, y_i^k \right) \right\}_{i=1}^{n_k},  
\]  
where \(\mathbf{x}_i^k \in \mathcal{X}\) represents sensor-derived features (e.g., accelerometer or gyroscope signals), \( y_i^k \in \mathcal{Y} \) denotes the activity label (e.g., walking, running), and \( n_k \) is the number of samples. Crucially, each source user’s data follows a distinct distribution \( P_k^{\text{Source}} \).  

For \( M \) unseen target users, the unlabeled datasets are given by:  
\[  
\mathcal{D}_m^{\text{Target}} = \left\{ \mathbf{x}_i^m \right\}_{i=1}^{n_m},  
\]  
where \(\mathbf{x}_i^m \in \mathcal{X}\) shares the same feature space as the source users, but the target distributions \( P_m^{\text{Target}} \) differ from source distributions (\( P_k^{\text{Source}} \neq P_m^{\text{Target}} \)).  

The goal is to train a model \( f: \mathcal{X} \rightarrow \mathcal{Y} \) using only \(\bigcup_{k=1}^K \mathcal{D}_k^{\text{Source}} \) to predict activity labels \(\{ y_i^m \}_{i=1}^{n_m}\) for all target users. The challenge lies in addressing distribution shifts (e.g., user-specific sensor placements, movement styles) while ensuring generalization to unseen domains.

\subsection{Sensor Position-Aware Graph Neural Network with Adversarial Domain Generalization}

\begin{figure}[h!]
\centering
\includegraphics[width=0.6\columnwidth]{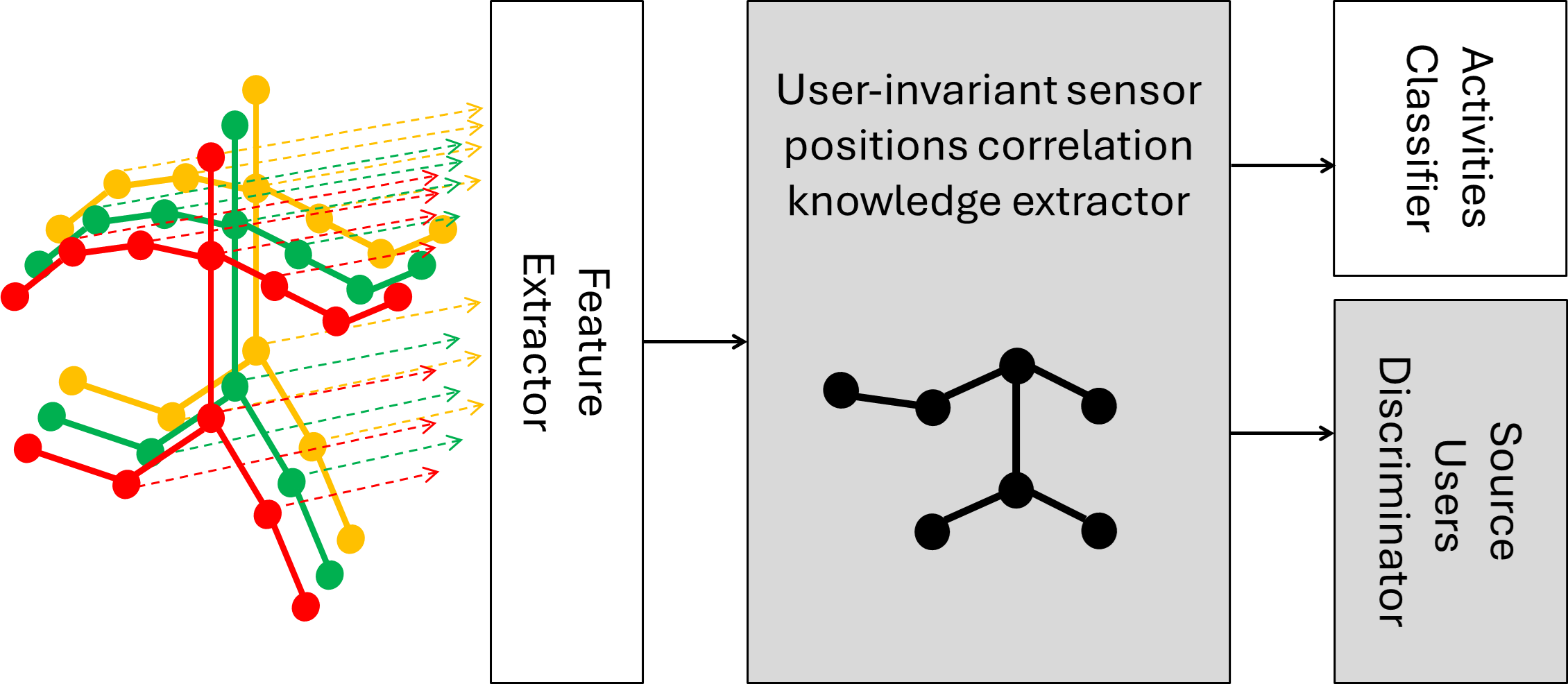}
\caption{Overview of the GNN-ADG architecture.\label{Overview_GNN-ADG_architecture}}
\end{figure}

Our proposed method, Sensor Position-Aware Graph Neural Network with Adversarial Domain Generalization (\textbf{GNN-ADG}), integrates GNNs with adversarial domain generalization to learn user-invariant features for effective cross-user activity recognition. The architecture consists of four main components, each playing a critical role in extracting meaningful features and ensuring robust generalization across users: \textbf{Feature Extractor}, \textbf{Sensor Positions Correlation Knowledge Extractor}, \textbf{Source Users Discriminator}, and \textbf{Activities Classifier}.

The shared and common knowledge of anatomical correlation between different body positions is explored for domain generalization in HAR. Capturing these intrinsic patterns of coordinated movement across different anatomical parts is crucial for improving the generalization capability of HAR models. An overview of the GNN-ADG architecture is illustrated in Figure~\ref{Overview_GNN-ADG_architecture}. Below, we provide a deeper discussion of the four components:

\begin{figure*}[h!] \centering \includegraphics[width=0.9\textwidth]{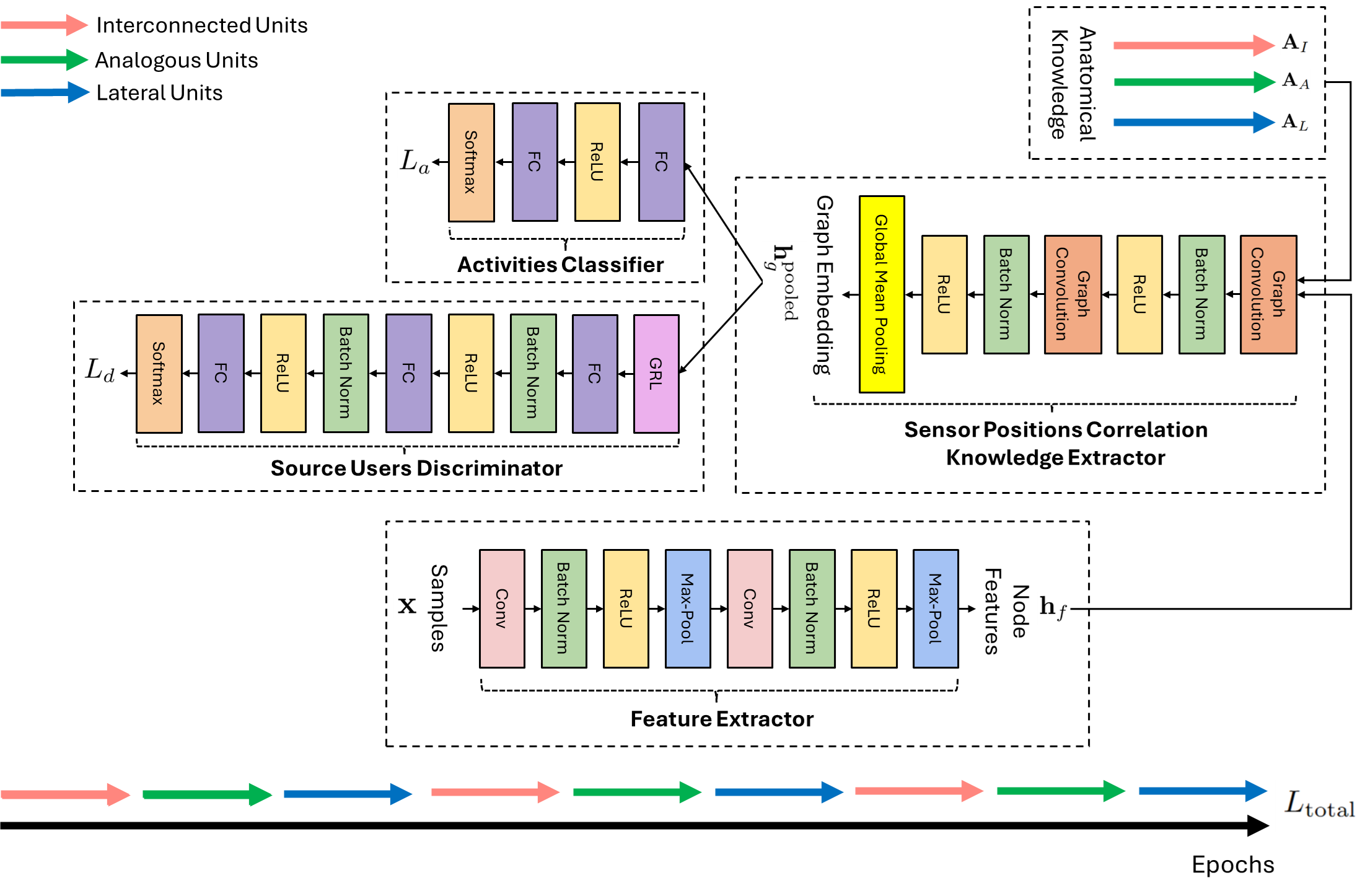} \caption{Information Fusion GNN-ADG framework overview.\label{GNN-ADG-no-edge-framework}} \end{figure*}

\subsubsection{Feature Extractor}

The Feature Extractor serves as the first hierarchy of common feature extraction, focusing on learning position-invariant features from the raw sensor signals. It extracts local patterns from the sensor data. This component processes the time-series data from multiple sensors placed on different body positions, such as the wrists, ankles, and core. This component ensures that low-level features are captured across various sensor placements.

The feature extractor acts as a generalized encoder applied uniformly across all sensor positions to learn common representations. These representations are independent of the specific location of the sensors on the body, providing a robust foundation for subsequent components. By treating sensors in different positions uniformly in this first layer, the model captures position-invariant signal features shared across users, improving generalization from the start.

We employ a CNN with two convolutional layers to extract the first hierarchy of common features as shown in Figure~\ref{GNN-ADG-no-edge-framework} Feature Extractor component. The first convolutional layer applies filters to capture short-term temporal patterns. The second convolutional layer builds upon the features from the first layer to capture more complex patterns. Both convolutional layers are followed by batch normalization, ReLU activation, and max-pooling to reduce dimensionality and mitigate overfitting.

Mathematically, the output of the feature extractor for input \( \mathbf{x} \) is:

\begin{equation}
    \mathbf{h}_f = \text{FeatureExtractor}(\mathbf{x})
\end{equation}

\subsubsection{Sensor Positions Correlation Knowledge Extractor}

The User-Invariant Sensor Positions Correlation Knowledge Extractor is the second hierarchy of common feature extraction and the highlight of our work. It focuses on modeling the relationships between different anatomical parts by leveraging GNNs. Sensors placed at different positions on the body correspond to distinct anatomical parts (e.g., wrists, knees, shoulders). The GNN models these spatial relationships to capture the intrinsic biomechanical correlations between these parts during various activities.

This component plays a crucial role in exploring and utilizing shared anatomical correlation knowledge across users. The coordinated movement patterns (e.g., the left knee moving forward as the right knee moves back during running) are modeled as a graph structure, where each node represents a sensor position, and the edges capture the anatomical correlation between them. This knowledge is user-independent and common across different individuals, despite variations in height, weight, gender, or personal movement styles.

By focusing on these common biomechanical patterns, the GNN learns user-invariant correlations between sensors, significantly enhancing the generalization capability of the model. This component ensures that the model can generalize effectively to unseen users by leveraging the common anatomical knowledge embedded in the relationships between different body parts.

In the design of this component, we utilize three distinct graph types, each capturing unique relationships between anatomical parts. These relationships allow the GNN to generalize movement patterns across users by capturing biomechanical correlations that are common across different body shapes, activity styles, and movement patterns. Each graph type contributes to an unique understanding of body mechanics, enhancing the model’s ability to interpret sensor data effectively.

\begin{figure*}[h!]
\centering
\includegraphics[width=0.8\textwidth]{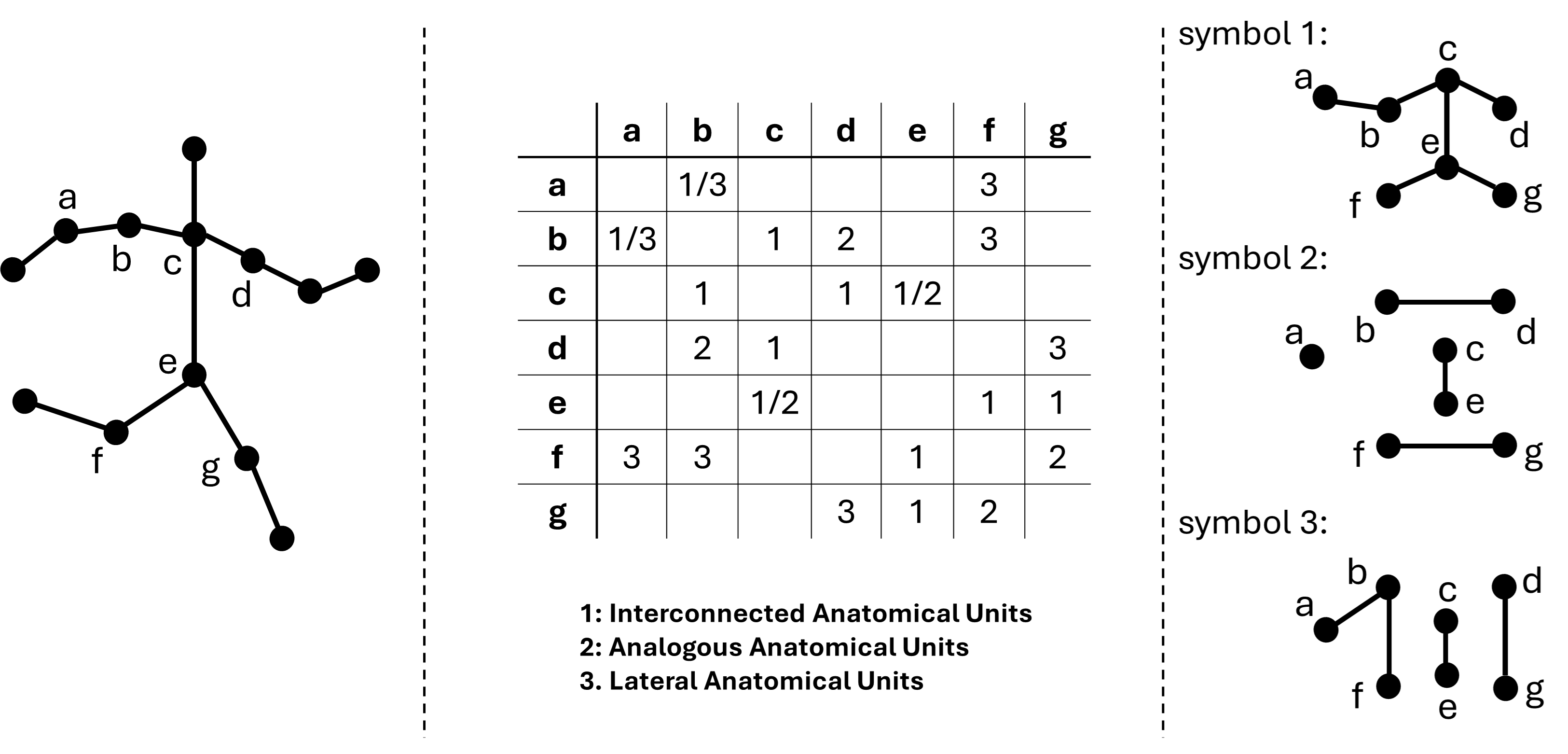}
\caption{Graph construction on common anatomical correlation of sensor position-aware knowledge across users.\label{3_categories_graphs}}
\end{figure*}

\paragraph{Graph Construction}

Figure~\ref{3_categories_graphs} outlines the graph construction methodology, which models common anatomical relations across users to capture common patterns in their data. On the left side of the figure, sensors attached to different parts of the human body are labeled as symbols \(a\), \(b\), \(c\), \(d\), \(e\), \(f\), and \(g\). The middle section of the figure shows the adjacency matrices corresponding to the three different graph types, represented by symbols 1, 2, and 3, which will be discussed in detail later. On the right side of the figure, these three types of anatomical correlations are visualized as distinct graphs, each capturing unique aspects of the shared anatomical relationships.

\textbf{Interconnected Anatomical Units}:
This graph type models spatially neighboring anatomical parts by connecting sensors located next to each other on the body. For example, sensors on the wrist and elbow would be linked as node a and b in Figure~\ref{3_categories_graphs} symbol 1. Neighboring parts tend to move in coordination due to physical proximity and biomechanical constraints. For instance, movements in the upper arm influence the forearm and wrist due to their connectedness. By linking spatial neighbors, this graph reflects the natural spatial connectivity and kinematic dependencies within the body. This graph provides the GNN with a basis for understanding immediate spatial dependencies, helping it to recognize local, neighboring movement correlations that are consistent across all users. This spatial adjacency graph ensures that the model accounts for primary, short-range anatomical dependencies in body movement.
        
\textbf{Analogous Anatomical Units}:
This graph type groups and connects sensors located on body parts with similar structures or functional purposes. For example, the left and right knees, shoulders, or wrists could be linked due to their mirrored roles in locomotion and balance as shown in Figure~\ref{3_categories_graphs} symbol 2. Symmetrical or functionally similar body parts often exhibit coordinated movements, especially in activities that require balance or mirrored motion (e.g., both knees bending simultaneously when squatting). By grouping these parts together, this graph enables the model to capture patterns based on bilateral or mirrored movement, which are common across human activities. This graph supports the GNN in recognizing functional symmetries, providing an understanding of how similar anatomical units work together. It enhances the model’s robustness by encoding correlations that are consistent due to the bilateral nature of the human body, aiding in activity recognition and generalization across users.

\textbf{Lateral Anatomical Units}:
This graph focuses on connecting sensors based on their position on the left, right, or middle side of the body. For example, it connects parts on the right side (e.g., right hand, right leg) as well as parts on the left side as shown in Figure~\ref{3_categories_graphs} symbol 3. Body parts positioned on the same side of the body often operate in coordination, especially for tasks that require dominant-side usage. For example, a person’s right hand and right leg may coordinate movements during activities like running, or may act together due to handedness or dominant-side preference. This graph connects body parts on the same side, capturing positional dependencies. By modeling side-specific movement patterns, the GNN gains a deeper understanding of coordinated actions across the left, right, and middle body parts. It is particularly useful for recognizing patterns related to handedness or dominant side usage, which contribute to the model’s generalization ability across users with varying preferences.

In summary, each of these graph types—capturing spatial correlation, functional correlation, and positional correlation—offers a unique perspective on human movement. By first exploring these graphs individually, we analyze their distinct contributions to capturing biomechanical relationships among sensor positions. \textbf{Interconnected Anatomical Units} emphasize local spatial dependencies, forming the foundation for detecting immediate movement correlations. \textbf{Analogous Anatomical Units} add a layer of understanding by grouping symmetrical or functionally similar anatomical parts, effectively capturing mirrored and coordinated movements. \textbf{Lateral Anatomical Units} highlight side-specific positional dependencies, enabling the model to detect asymmetrical or dominant-side activities.

\paragraph{Graph Neural Network}

The extracted features \( \mathbf{h}_f \) from the feature extractor combined with anatomical correlation knowledge (i.e. three types of graphs discussed in previous section) as shown in Figure~\ref{GNN-ADG-no-edge-framework} are passed through a GNN to model the interactions between sensor positions. Specifically, we employ Graph Convolutional Networks (GCNs) \cite{Kipf:2016tc} with two graph convolutional layers, as depicted in the Sensor Positions Correlation Knowledge Extractor component. 

Each graph convolutional layer is followed by batch normalization and a ReLU activation function to enhance stability and introduce non-linearity. At the final stage, a global mean pooling operation is applied to aggregate the learned position-aware embeddings across all nodes into a single feature vector, summarizing the graph's information.

\begin{enumerate}
    \item First Graph Convolutional Layer:
    \begin{equation}
        \mathbf{h}_g^{(1)} = \sigma\left(\text{BatchNorm}\left( \hat{\mathbf{A}} \mathbf{h}_f \mathbf{W}^{(1)} \right)\right)
    \end{equation}
    \item Second Graph Convolutional Layer:
    \begin{equation}
        \mathbf{h}_g^{(2)} = \sigma\left(\text{BatchNorm}\left( \hat{\mathbf{A}} \mathbf{h}_g^{(1)} \mathbf{W}^{(2)} \right)\right)
    \end{equation}
\end{enumerate}

Here:
\begin{itemize}
    \item \( \hat{\mathbf{A}} = \mathbf{D}^{-1/2} \mathbf{A} \mathbf{D}^{-1/2} \) is the normalized adjacency matrix, where \( \mathbf{A} \) represents the adjacency matrix of the graph and \( \mathbf{D} \) is the degree matrix. We define three distinct adjacency matrices $\mathbf{A}_I$, $\mathbf{A}_A$, and $\mathbf{A}_L$ corresponding to Interconnected, Analogous, and Lateral Units respectively.
    \item \( \mathbf{W}^{(1)} \) and \( \mathbf{W}^{(2)} \) are the learnable weight matrices for the first and second convolutional layers, respectively.
    \item \( \sigma \) is the ReLU activation function, applied to introduce non-linearity.
    \item \(\text{BatchNorm}(\cdot)\) denotes batch normalization applied to the output of each convolution layer.
\end{itemize}

The sensor positions are modeled as nodes in a graph \( G = (V, E) \), where \( V \) denotes the set of nodes (sensor positions) and \( E \) denotes the set of edges, representing the relationships between sensors. These edges are pre-defined based on prior knowledge of the sensor layout, as shown in Figure~\ref{3_categories_graphs}, and capture interconnected anatomical units, analogous anatomical units, and lateral anatomical units.

After the final convolutional layer, a global mean pooling operation is applied:
\begin{equation}
    \mathbf{h}_g^{\text{pooled}} = \text{GlobalMeanPooling}\left(\mathbf{h}_g^{(2)}\right)
\end{equation}

This operation aggregates the learned embeddings from all nodes into a single feature vector, summarizing the graph's information in a compact form. The resulting embeddings incorporate both local and global sensor relationships, enabling robust modeling of the common biomechanical correlations of different body positions across users.

\subsubsection{Activities Classifier}
The Activities Classifier $C_a$ is responsible for performing the HAR task. It takes the graph embedding features $\mathbf{h}_g^{\text{pooled}}$ and identifies the specific activity being performed (e.g., walking, running, sitting) as shown in Figure~\ref{GNN-ADG-no-edge-framework}. The classifier is trained to map the extracted user-invariant features to the corresponding activity labels. This ensures that the model not only learns user-agnostic features but also achieves high performance in activity recognition across different users.

The classifier focuses solely on the HAR task, ensuring that the features learned by the upstream components are accurately translated into activity predictions. By relying on features that capture the common anatomical correlations knowledge cross users, the classifier achieves robust recognition across users.

The classifier predicts activity labels \(\hat{y}\) through:
\begin{equation}
    \hat{y} = C_a\left(\mathbf{h}_g^{\text{pooled}}\right)
\end{equation}

with the \textit{Activity Classification Loss} enforcing label alignment based on the predicted activity labels \( \hat{y} \) and the true labels \( y \):
    \begin{equation}
        L_a = \text{CrossEntropy}(y, \hat{y})
    \end{equation}

\subsubsection{Source Users Discriminator}

This component is designed to ensure that the learned features are domain-invariant, meaning they generalize effectively across different users. To achieve this, it leverages a GRL to facilitate adversarial learning. The GRL plays a critical role by reversing the gradient direction during learning, enabling the feature extractor and the user-invariant sensor positions correlation knowledge extractor to learn user-independent features.

During the discrimination phase, the Source Users Discriminator attempts to predict the user domain (i.e., the specific user label associated with the input features) based on outputs $\mathbf{h}_g^{\text{pooled}}$ from the Sensor Positions Correlation Knowledge Extractor. During the confusion phase, the GRL reverses the gradients, thereby pushing the Feature Extractor and Sensor Positions Correlation Knowledge Extractor to learn features that make it difficult for the Source Users Discriminator to identify the user domains. As a result, these components learn user-invariant features that are not biased toward any specific user or domain.

This adversarial process operates in a push-and-pull manner, driving the model to capture shared, user-independent feature representations. The GRL pushes the model to suppress user-specific characteristics by reversing the gradients of the domain discriminator, while the domain discriminator pulls the representations toward domain-distinguishable patterns. This dynamic interaction balances the competing objectives, encouraging the model to learn representations that is user-agnostic. This mechanism enhances the model’s ability to generalize effectively across different users, achieving robust performance on unseen users without requiring additional calibration or fine-tuning.

To achieve that, we introduce the Source Users Discriminator \( D \), which aims to predict the user domain \( \hat{d} \) based on the extracted graph embedding features $\mathbf{h}_g^{\text{pooled}}$. The domain discriminator is trained adversarially using the GRL. The domain discriminator \( D \) predicts the user domain labels \( \hat{d} \) based on the transformed feature embeddings:

\begin{equation}
    \hat{d} = D(\text{GRL}(\mathbf{h}_g^{\text{pooled}}))
\end{equation}

This adversarial dynamic is formalized through the \textit{Source Users Discriminator Loss}:
    \begin{equation}
        L_d = \text{CrossEntropy}(d, \hat{d})
    \end{equation}

This loss encourages the model to produce domain-invariant features by predicting the user domain \( \hat{d} \) based on the true user domain labels \( d \). It helps the model generalize across different users.

In summary, by combining the Feature Extractor, Sensor Positions Correlation knowledge Extractor, and adversarial learning in the Source Users Discriminator, we create a model capable of learning user-invariant representations. This design enables robust cross-user generalization, improving the model’s performance on new, unseen users without requiring domain-specific adjustments.

\subsubsection{Unified Graph with Cyclic Strategy}

To fully leverage the complementary strengths of the three graph types—Interconnected, Analogous, and Lateral Units—we introduce a unified graph Information Fusion GNN-ADG approach paired with a cyclic training strategy. Rather than training separate models for each graph type, this method integrates all three edge topologies into a single model and systematically cycles through them during training. During training, the model cycles through these matrices at intervals of $N$ epochs. At epoch $t$, the active adjacency matrix $\mathbf{A}^{(t)}$ is determined by:
\begin{equation}
\mathbf{A}^{(t)} = \mathbf{A}_{\tau\left( \left\lfloor \frac{t}{N} \right\rfloor \bmod 3 \right)},
\end{equation}
where $\tau: \mathbb{Z} \to {I, A, L}$ maps the cyclic phase such that:
\[
\tau(k) = 
\begin{cases}
I \text{(Interconnected Units)} & \text{if } k \bmod 3 = 0, \\
A \text{(Analogous Units)} & \text{if } k \bmod 3 = 1, \\
L \text{(Lateral Units)} & \text{if } k \bmod 3 = 2.
\end{cases}
\]
This ensures:

$\mathbf{A}_I$ is used for epochs $t \in [0, N-1]$,
$\mathbf{A}_A$ for $t \in [N, 2N-1]$,
$\mathbf{A}_L$ for $t \in [2N, 3N-1]$, with the cycle repeating every $3N$ epochs. For instance, if $N=20$, the pattern restarts at $t=60$, mirroring the original 20-epoch interval design. This cycling allows the GNN to refine its understanding of inter-sensor relationships by alternately emphasizing different anatomical perspectives, fostering a dynamic and holistic learning process.

The fundamental motivation behind this approach stems from the inherent complexity of human movement patterns. Different activities fundamentally rely on distinct biomechanical relationships—walking emphasizes the coordination between opposite limbs captured by Analogous Units, while unilateral movements like reaching utilize connections represented in Lateral Units. By employing a static graph structure, the model would inevitably favor certain movement correlations over others, potentially compromising its generalization capacity across the full spectrum of human activities and individual movement styles. Our cyclic approach addresses this limitation by ensuring the model systematically learns from all relevant anatomical relationships.
During each training phase, the model optimizes a phase-specific loss function that depends on the current graph topology, where for each graph type $k \in {I, A, L}$, the corresponding loss function is:
\begin{equation}
L_{k} = L_a^{(k)} + \beta L_d^{(k)}
\end{equation}
Here, $L_a^{(k)}$ represents the activity classification loss and $L_d^{(k)}$ represents the source users discriminator loss, both computed using the adjacency matrix $\mathbf{A}_k$. The hyperparameter $\beta$ balances the contributions of these two loss components, controlling the trade-off between optimizing for activity classification accuracy and ensuring domain invariance across different users.
Over the complete training process, the model effectively optimizes a combined objective that integrates insights from all graph perspectives:
\begin{equation}
L_{\text{total}} = \sum_{k \in {I,A,L}} L_k
\end{equation}
This formulation ensures that the final model parameters perform well across all graph topologies, rather than specializing to any single perspective. Our cyclic strategy, where the GNN is systematically exposed to progressively complex and diverse graph types, ensures comprehensive adaptation to the full spectrum of movement correlations. By alternating focus among the graph types, it prevents overfitting to a single correlation type and encourages the learning of features that are invariant across anatomical perspectives. This results in a robust, user-invariant representation capable of handling diverse activities and unseen users effectively.

\section{Experimental Setup}
\label{sec:experimental_setup}

To evaluate the effectiveness of our proposed \textbf{GNN-ADG} method for cross-user activity recognition, we conduct experiments on two widely-used benchmark datasets: \textit{OPPORTUNITY} (OPPT) and \textit{Daily and Sports Activities Dataset} (DSADS). These datasets provide a diverse set of activities and sensor configurations, making them ideal for assessing the generalization capabilities of our approach. A summary of the subjects clusters, common activities across users and the sensor positions of the two datasets is provided in Table~\ref{tab_datasets_info}.

\begin{table}[h!]
\caption{Two sensor-based HAR datasets information.}
\label{tab_datasets_info}
\centering
\scriptsize 
\resizebox{\columnwidth}{!}{%
\begin{tabular}{|p{1.0cm}|p{1.5cm}|p{8.0cm}|p{2.6cm}|}
\hline
\textbf{Dataset} & \textbf{Domains} & \textbf{Common Activities} & \textbf{Sensor Positions} \\ \hline
OPPT & \begin{tabular}[c]{@{}l@{}}A = [S1], \\
B = [S2],\\
C = [S3],\\
D = [S4]\end{tabular} & \begin{tabular}[c]{@{}l@{}}0: Open Door 1, 1: Open Door 2, 2: Close Door 1,\\
3: Close Door 2, 4: Open Fridge, 5: Close Fridge,\\
6: Open Dishwasher, 7: Close Dishwasher, 8: Open Drawer 1,\\
9: Close Drawer 1, 10: Open Drawer 2, 11: Close Drawer 2,\\
12: Open Drawer 3, 13: Close Drawer 3, 14: Clean Table,\\
15: Drink From Cup, 16: Toggle Switch\end{tabular} &\begin{tabular}[c]{@{}l@{}} 0: Back,\\ 1: Right Upper Arm,\\ 2: Right Lower Arm,\\ 3: Left Upper Arm,\\ 4: Left Lower Arm \end{tabular}\\ \hline

DSADS & \begin{tabular}[c]{@{}l@{}}A = [1,2],\\
B = [3,4],\\
C = [5,6],\\
D = [7,8]\end{tabular} & \begin{tabular}[c]{@{}l@{}}0: Sitting, 1: Standing, 2: Lying On Back, \\
3: Lying On Right, 4: Ascending Stairs, \\
5: Descending Stairs, 6: Standing In Elevator Still, \\
7: Moving Around In Elevator, \\
8: Walking In Parking Lot, \\
9: Walking On Treadmill In Flat, \\
10: Walking On Treadmill Inclined Positions, \\
11: Running On Treadmill In Flat, \\
12: Exercising On Stepper, \\
13: Exercising On Cross Trainer, \\
14: Cycling On Exercise Bike In Horizontal Positions, \\
15: Cycling On Exercise Bike In Vertical Positions, \\
16: Rowing, 17: Jumping, 18: Playing Basketball\end{tabular} & \begin{tabular}[c]{@{}l@{}}0: Torso,\\ 1: Right Arm,\\ 2: Left Arm,\\ 3: Right Leg,\\ 4: Left Leg\end{tabular} \\ \hline
\end{tabular}%
}
\end{table}

\subsection{Datasets and Experimental setup}

\textbf{OPPT} \cite{chavarriaga2013opportunity} is a dataset that captures daily living activities using multiple body-worn sensors. It includes data from four subjects performing activities such as walking, standing, sitting, and object interaction. The dataset is particularly challenging due to the presence of fine-grained activities that involve highly similar motions, such as opening/closing different drawers (e.g., Drawer 1, 2, 3) or doors, where the distinction lies only in the specific location or layer interacted with. These subtle differences require models to discern minor variations in sensor data. The dataset contains a rich set of sensory data from accelerometers, gyroscopes, and magnetometers placed on various body parts.

\textbf{DSADS} \cite{barshan2014recognizing} consists of data collected from eight subjects performing 19 activities, including everyday and sports activities. Each subject wore inertial measurement units on five different body parts, providing comprehensive motion data. While the activities are diverse (e.g., cycling, rowing, playing basketball), they generally involve more distinct motion patterns compared to OPPORTUNITY, making classification relatively less ambiguous.

We adopt a leave-one-subject-out cross-validation protocol to simulate a cross-user scenario. For each dataset, the model is trained on data from all subject clusters except one, and then tested on the left-out cluster. This approach ensures that the evaluation is conducted on entirely unseen users, offering a robust assessment of the model's generalization capability.

\subsection{Cross-user HAR performance results}

To demonstrate the effectiveness of our proposed method, we compare \textbf{GNN-ADG} and \textbf{Information Fusion GNN-ADG} with the following state-of-the-art domain generalization methods:

\begin{itemize}
    \item \textbf{ERM} \cite{zhang2018mixup}: Empirical Risk Minimization (ERM) is a foundational approach that aims to minimize the average loss over all training examples. It serves as a baseline for domain generalization and focuses on learning from the data without explicitly accounting for domain-specific variations.
    \item \textbf{RSC} \cite{huang2020self}: Representation Self-Challenging (RSC) improves cross-domain generalization by discarding dominant features during training and forcing the network to use other label-related features. This method is simple, effective, and requires no prior knowledge of new domains or additional network parameters.

    \item \textbf{ANDMask} \cite{parascandolo2021learning}: Focuses on learning invariant features by enforcing updates that improve performance across all training domains simultaneously. This method uses a logical AND approach to emphasize invariances and prevent memorization, outperforming traditional regularizers in tasks with clear distinctions between invariant and spurious features.

    \item \textbf{AdaRNN} \cite{du2021adarnn}: Addresses the Temporal Covariate Shift problem in time series data, where distribution changes over time hinder forecasting accuracy. AdaRNN employs two key algorithms: Temporal Distribution Characterization to capture distribution information and Temporal Distribution Matching to reduce mismatch, enabling adaptive and generalizable models.

    \item \textbf{DIFEX} \cite{lu2022domain}: Proposes Domain-Invariant Feature Exploration (DIFEX) to enhance domain generalization by combining internal and mutual invariance. Internal invariance focuses on intrinsic features within a domain, while mutual invariance captures transferable features across domains.
\end{itemize}

\begin{figure*}[h!] \centering \includegraphics[width=1.0\textwidth]{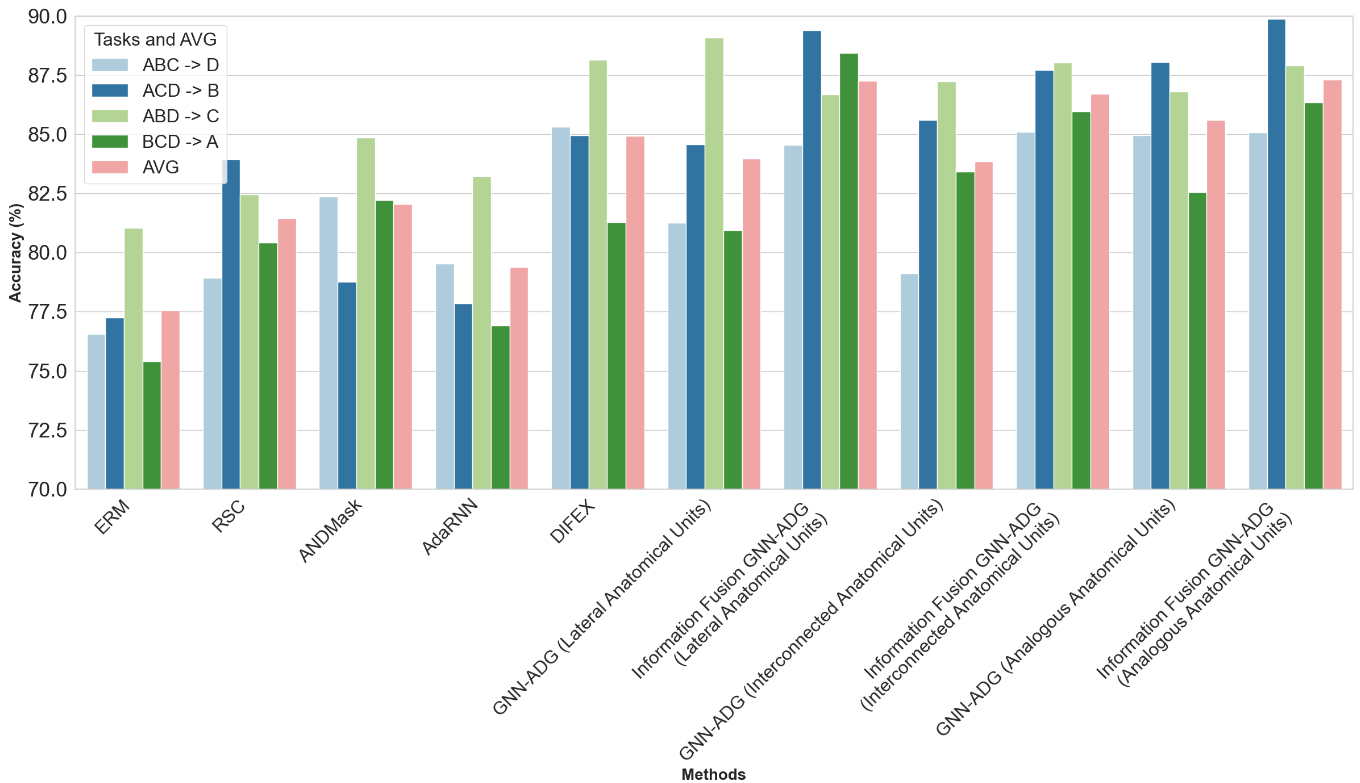} \caption{DSADS Dataset Results.\label{DSADS_Dataset_Result}} \end{figure*}

\begin{figure*}[h!] \centering \includegraphics[width=1.0\textwidth]{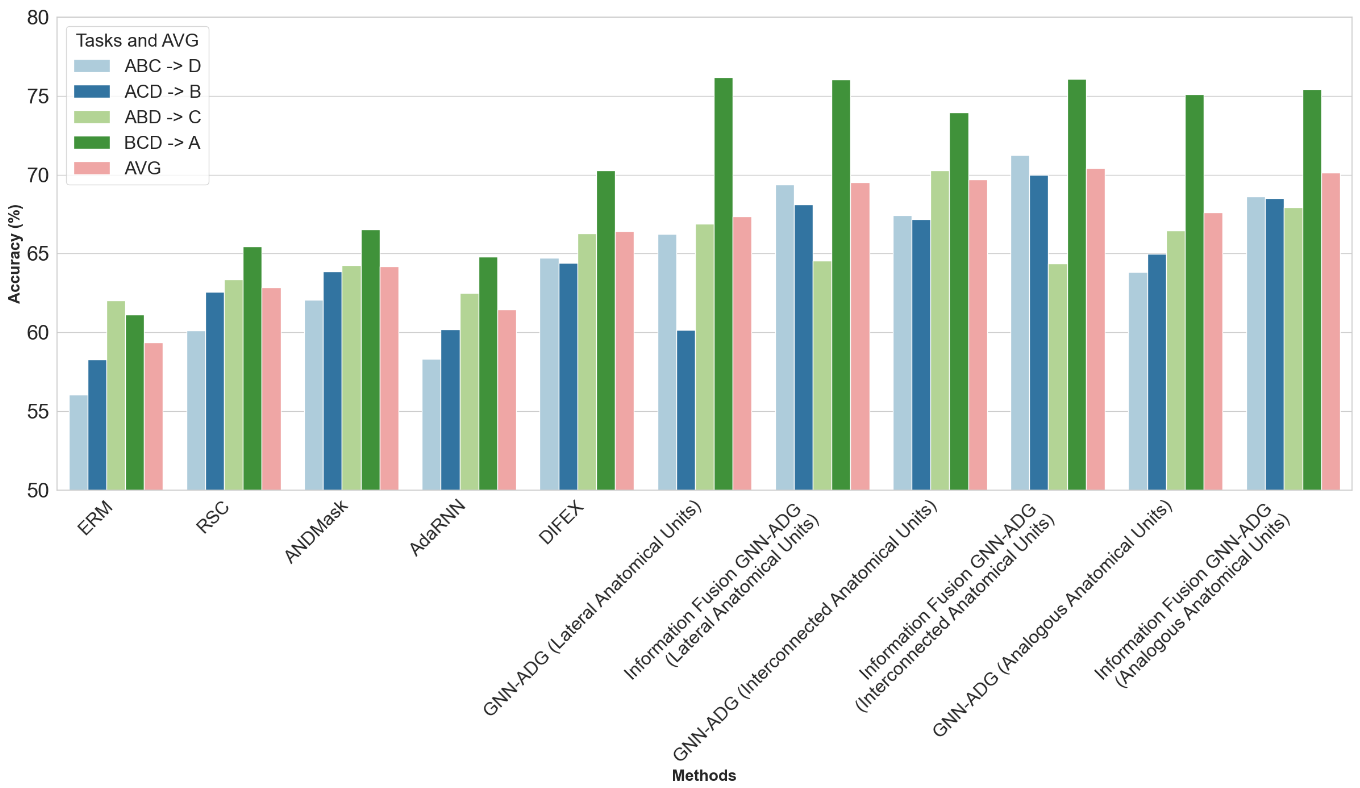} \caption{OPPT Dataset Results.\label{OPPT_Dataset_Result}} \end{figure*}

These methods have been selected for their relevance and performance in domain generalization tasks. By comparing against these baselines, we aim to demonstrate the advantages of our GNN-ADG and Information Fusion GNN-ADG methods in handling cross-user variability without requiring access to target domain data during training. To evaluate the effectiveness of our proposed methods, we conducted experiments on two benchmark datasets: DSADS and OPPT. We compared GNN-ADG and Information Fusion GNN-ADG—each with three anatomical relations—against state-of-the-art domain generalization methods, including ERM, RSC, ANDMask, AdaRNN, and DIFEX. The results are presented in Table \ref{tab:comparison_DSADS}, Figure \ref{DSADS_Dataset_Result}, Table \ref{tab:comparison_OPPT}, and Figure \ref{OPPT_Dataset_Result}.

\begin{table}[ht]
\centering
\caption{Comparison of methods with average accuracy and standard deviation values on DSADS dataset.}
\begin{tabular}{lcc}
\toprule
\textbf{Method} & \textbf{Average} & \textbf{Standard deviation} \\
\midrule
ERM & 77.56 & 2.44 \\
RSC & 81.44 & 2.21 \\
ANDMask & 82.05 & 2.51 \\
AdaRNN & 79.38 & 2.78 \\
DIFEX & 84.93 & 2.82 \\
GNN-ADG (Lateral Anatomical Units) & 83.97 & 3.79 \\
Information Fusion GNN-ADG (Lateral Anatomical Units) & 87.27 & 2.13 \\
GNN-ADG (Interconnected Anatomical Units) & 83.85 & 3.52 \\
Information Fusion GNN-ADG (Interconnected Anatomical Units) & 86.70 & \textbf{1.41} \\
GNN-ADG (Analogous Anatomical Units) & 85.59 & 2.39 \\
Information Fusion GNN-ADG (Analogous Anatomical Units) & \textbf{87.31} & 2.06 \\
\bottomrule
\end{tabular}
\label{tab:comparison_DSADS}
\end{table}

\textbf{Methods Performance Analysis.} DSADS includes a variety of activities involving both static postures (e.g., sitting, standing) and dynamic movements (e.g., walking, running). Sensors are placed on the torso, arms, and legs to capture full-body dynamics. As shown in Table \ref{tab:comparison_DSADS}, among baseline methods, DIFEX achieves the highest performance (84.93\% average accuracy) by promoting diversity and aligning domain-invariant features. However, it fails to explicitly model the interdependencies between different anatomical parts, limiting its ability to generalize in complex full-body activities such as "cycling" and "rowing."
Our GNN-ADG approach with all three types of anatomical knowledge demonstrates competitive or superior performance compared to DIFEX. GNN-ADG with Analogous Anatomical Units achieves a higher accuracy (85.59\%) than DIFEX, while GNN-ADG with Lateral Anatomical Units (83.97\%) and Interconnected Anatomical Units (83.85\%) show comparable performance. This indicates that incorporating explicit anatomical relationships through graph neural networks effectively captures the biomechanical patterns necessary for cross-user generalization, particularly for full-body activities where coordinated movements are crucial. As illustrated in Figure \ref{DSADS_Dataset_Result}, GNN-ADG with Analogous Anatomical Units performs especially well for activities involving bilateral or symmetrical movements, such as "jumping" or "rowing," where both arms or legs work together.

\begin{table}[ht]
\centering
\caption{Comparison of methods with average accuracy and standard deviation values on OPPT dataset.}
\begin{tabular}{lcc}
\toprule
\textbf{Method} & \textbf{Average} & \textbf{Standard deviation} \\
\midrule
ERM & 59.37 & 2.73 \\
RSC & 62.87 & 2.20 \\
ANDMask & 64.17 & \textbf{1.84} \\
AdaRNN & 61.44 & 2.81 \\
DIFEX & 66.43 & 2.69 \\
GNN-ADG (Lateral Anatomical Units) & 67.37 & 6.62 \\
Information Fusion GNN-ADG (Lateral Anatomical Units) & 70.53 & 3.71 \\
GNN-ADG (Interconnected Anatomical Units) & 69.70 & 3.16 \\
Information Fusion GNN-ADG (Interconnected Anatomical Units) & \textbf{70.95} & 3.98 \\
GNN-ADG (Analogous Anatomical Units) & 67.59 & 5.11 \\
Information Fusion GNN-ADG (Analogous Anatomical Units) & 70.12 & 3.56 \\
\bottomrule
\end{tabular}
\label{tab:comparison_OPPT}
\end{table}

OPPT focuses on upper-body activities such as opening/closing doors and drawers, toggling switches, and cleaning tasks. Sensors are positioned on the back, upper arms, and lower arms, capturing localized upper-body dynamics. As presented in Table \ref{tab:comparison_OPPT}, DIFEX achieves the highest performance among baseline methods (66.43\%), highlighting its generalization ability across users. However, as evident from Figure \ref{OPPT_Dataset_Result}, it lacks the ability to model localized correlations between neighboring body parts, which are essential for OPPT tasks.
Our GNN-ADG variants all outperform DIFEX on this dataset. GNN-ADG with Interconnected Anatomical Units achieves 69.70\% accuracy, surpassing DIFEX by 3.27\%, while GNN-ADG with Analogous Anatomical Units (67.59\%) and Lateral Anatomical Units (67.37\%) also show improvements. The superior performance of GNN-ADG with Interconnected Anatomical Units is particularly noteworthy because OPPT tasks heavily rely on localized coordination between adjacent body parts. For example, during "cleaning a table," the upper arm and lower arm sensors must work together, and modeling their spatial dependencies directly enhances performance.

\textbf{Information Fusion Analysis.} When examining the impact of information fusion, we observe improvements across all anatomical graph types in both datasets, as shown in Tables \ref{tab:comparison_DSADS} and \ref{tab:comparison_OPPT}. For the DSADS dataset, Information Fusion GNN-ADG with Lateral Anatomical Units shows a substantial improvement of 3.30\% (from 83.97\% to 87.27\%), Interconnected Anatomical Units demonstrate a 2.85\% improvement (from 83.85\% to 86.70\%) with the most significant reduction in standard deviation (from 3.52\% to 1.41\%), and Analogous Anatomical Units show a 1.72\% improvement (from 85.59\% to 87.31\%), achieving the highest overall performance.
For the OPPT dataset, Information Fusion GNN-ADG with Lateral Anatomical Units shows a 3.16\% improvement (from 67.37\% to 70.53\%) and substantial reduction in standard deviation (from 6.62\% to 3.71\%), Interconnected Anatomical Units achieve a 1.25\% improvement (from 69.70\% to 70.95\%), reaching the highest overall performance, and Analogous Anatomical Units demonstrate a 2.53\% improvement (from 67.59\% to 70.12\%).

These results indicate that different anatomical graph types benefit from information fusion to varying degrees depending on the nature of the activities. For full-body activities in DSADS, Lateral and Interconnected Units benefit more significantly from information fusion than Analogous Units. This can be attributed to the complementary nature of the anatomical knowledge they capture. When fused, these provide a more complete biomechanical representation of the activity. For the OPPT dataset, Lateral Units benefit most from information fusion. This can be explained by the nature of upper-body tasks like opening doors or cleaning, which may rely heavily on coordinated movements between both arms. The fusion of lateral relationships with the already strong interconnected relationships creates a more comprehensive representation of upper-body biomechanics.

Comparing the patterns across both datasets reveals that Interconnected Anatomical Units show the most consistent performance across both datasets, highlighting the fundamental importance of modeling direct anatomical connections regardless of activity type. Analogous Anatomical Units perform exceptionally well for full-body activities (DSADS) but less so for localized upper-body tasks (OPPT), indicating their particular strength in capturing coordinated bilateral movements common in activities like walking or jumping. Lateral Anatomical Units benefit substantially from information fusion in both datasets, suggesting they provide complementary information that enhances the other graph types but may be insufficient when used alone.

This demonstrates that different types of anatomical relationships contribute uniquely to activity recognition, and their unified representation through information fusion creates a more robust and generalizable model for cross-user HAR. Moreover, adversarial learning is facilitated in GNN-ADG by a GRL, which is a crucial component for extracting user-invariant features. The GRL works by reversing the gradients of the domain discriminator during backpropagation, enabling the model to learn representations that confuse the Source Users Discriminator. This setup encourages the model to produce features that are indistinguishable across different user domains, effectively making the features domain-invariant.

The results across both datasets highlight the unique strengths of our proposed approach. By combining Graph Neural Networks with adversarial learning and leveraging the complementary information from different anatomical graph structures through the information fusion mechanism, GNN-ADG captures the intricate interplay of biomechanical relationships, enabling robust cross-user HAR performance. These findings emphasize the importance of bridging biomechanical insights with machine learning models to tackle domain generalization challenges in cross-user sensor-based HAR.

\subsection{Model Interpretability Analysis}

\begin{figure}[h!]
\centering
\includegraphics[width=1.0\columnwidth]{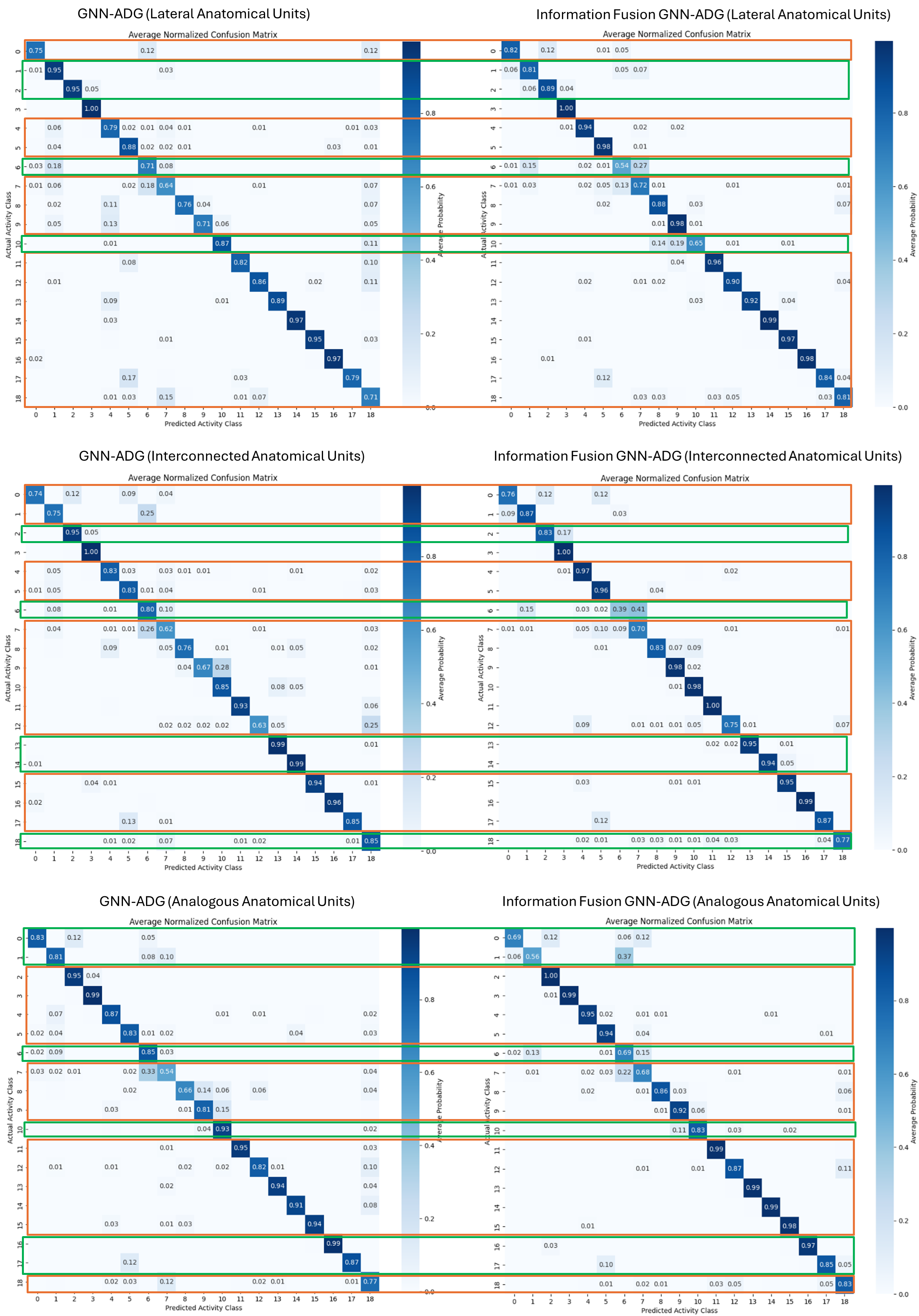}
\caption{Comparison of Accuracy and Confusion Matrix: GNN-ADG vs. Information Fusion GNN-ADG on the DSADS Dataset.\label{dsads_cm_compare}}
\end{figure}

In this section, we conduct a model interpretability analysis to compare the performance of GNN-ADG (using single anatomical knowledge) and Information Fusion GNN-ADG (integrating all three anatomical knowledge types). While the unified graph with cyclic strategy improves overall performance across all graph topologies, we further investigate which specific activities benefit most from this fusion mechanism. Figure~\ref{dsads_cm_compare} compares the accuracy and confusion matrices of GNN-ADG and Information Fusion GNN-ADG on the DSADS dataset. The orange rectangles highlight activities with improved performance after fusion, while the green rectangles indicate those with declined performance. 

Activities that leverage all three anatomical perspectives—Interconnected, Analogous, and Lateral—show the greatest improvement due to their reliance on a holistic integration of movement patterns. These include: Ascending stairs (4), Descending stairs (5), Moving around in an elevator (7), Walking in a parking lot (8), Walking on a treadmill in flat positions (9), Running on a treadmill in flat positions (11), Exercising on a stepper (12), Cycling on an exercise bike in vertical positions (15). For example, ascending stairs (4) and descending stairs (5) require coordination between legs (Interconnected), symmetry between left and right limbs (Analogous), and unilateral leg strength (Lateral). The cyclic training strategy alternates between these perspectives, effectively capturing these relationships and enhancing recognition. Similarly, locomotion activities like walking in a parking lot (8), walking on a treadmill in flat positions (9), and running on a treadmill in flat positions (11) involve rhythmic limb coordination, symmetry, and side-specific adjustments. Equipment-based tasks, such as exercising on a stepper (12) and cycling on an exercise bike in vertical positions (15), demand synchronized movements tailored to dynamic interactions with equipment. By preventing overfitting to a single anatomical perspective, this approach improves generalization, making it adept at recognizing these complex activities.

Activities benefiting from two of the three unit types still see improvements, though less pronounced, as they involve a mix of movement patterns addressed by a subset of anatomical graphs. These include: Sitting (0), Exercising on a cross trainer (13), Cycling on an exercise bike in horizontal positions (14), Rowing (16), Jumping (17), Playing basketball (18). For instance, playing basketball (18), enhanced by Lateral and Analogous units, involves varied movements like running, jumping, and lateral shifts. The fusion of these two perspectives improves recognition of these complex patterns, even without the full contribution of the Interconnected unit. These activities benefit from the model, but their reliance on fewer biomechanical relationships limits the extent of improvement compared to those utilizing all three units.

Static or less complex activities improved by only one unit type show limited gains from the fusion approach. These include: Standing (1) (Interconnected), Lying on back (2) (Analogous), Lying on the right (3) (Analogous), Walking on a treadmill in inclined positions (10) (Interconnected). These tasks involve minimal movement or specific biomechanical relationships, making the additional complexity of the fusion model less advantageous. For example, standing (1) and lying on back (2) are static, while walking on a treadmill in inclined positions (10) relies primarily on leg coordination (Interconnected). The extra perspectives may introduce noise rather than insight, reducing the model’s effectiveness for these activities. In fact, Information Fusion GNN-ADG, which is designed to capture diverse biomechanical relationships through interconnected units. It excels at activities with rich, coordinated movements but might overcomplicate something simple like standing still. Here, the model’s focus on dynamic interactions becomes a weakness, as it introduces unnecessary complexity for static tasks, resulting in degraded performance.

\begin{figure}[h!]
\centering
\includegraphics[width=1.0\columnwidth]{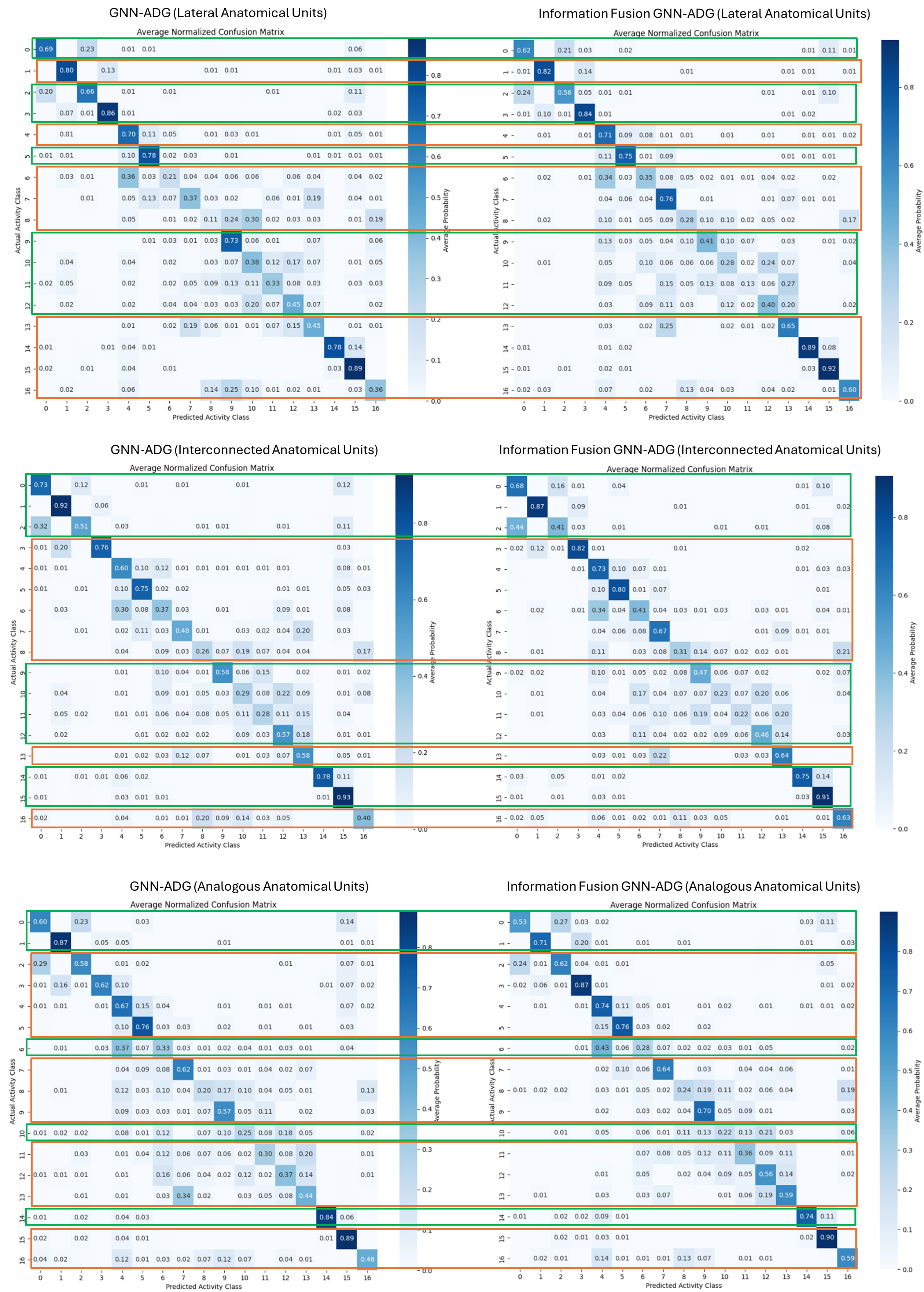}
\caption{Comparison of Accuracy and Confusion Matrix: GNN-ADG vs. Information Fusion GNN-ADG on the OPPT Dataset.\label{oppt_cm_compare}}
\end{figure}

Figure~\ref{oppt_cm_compare} compares the accuracy and confusion matrices of GNN-ADG and Information Fusion GNN-ADG on the OPPT dataset. The OPPT dataset is designed to capture upper-body movements that emphasizes the arms and torso. The fusion model demonstrates improved performance for activities that engage a combination of unilateral actions, coordinated movements, and symmetric postures, as these tasks leverage the strengths of the three anatomical unit types. Specifically, activities such as Close Door 2 (3), Open Fridge (4), Close Dishwasher (7), Open Drawer 1 (8), Close Drawer 3 (13), Drink From Cup (15), and Toggle Switch (16) exhibit the most significant improvement. For example, opening a fridge likely involves a unilateral arm movement to pull the door (lateral), coordination between the arm and back to maintain posture (interconnected), and symmetric engagement of the upper body for balance (analogous). Similarly, toggling a switch requires fine motor control with one hand (lateral), precise coordination between the arm and torso (interconnected), and habitual symmetry in posture (analogous). The placement of sensors on the back and both arms enables the model to capture these multifaceted interactions effectively. By alternating between the lateral, interconnected, and analogous perspectives during training, the model avoids overfitting to a single movement pattern, instead developing a holistic understanding of these complex upper-body activities. This adaptability makes it adept at recognizing tasks that require diverse contributions from the torso and arms, aligning seamlessly with the OPPT dataset’s upper-body focus.

Activities that show improvement from only one unit type or decline across all three, such as Open Door 1 (0) and Open Drawer 2 (10), highlight limitations of the Information Fusion GNN-ADG model within the OPPT dataset. These tasks exhibit movement patterns that do not align well with the generalized anatomical relationships defined by the lateral, interconnected, and analogous units. For instance, opening Door 1 might involve a specific arm angle or body position—such as a twist or reach not typical of other door-opening actions—that the sensors on the back and arms fail to capture in a way that fits the model’s framework. Moreover, the OPPT dataset includes activities that are highly similar, such as different instances of door-opening or drawer-opening, which may involve nearly identical movement patterns. For example, Open Door 1 (0) and Open Door 2 (1) might differ only in minor contextual factors, like door height or angle, while opening various drawers—such as Open Drawer 1 (8), Open Drawer 2 (10), and Open Drawer 3 (12)—could require comparable unilateral arm extensions. In such cases, the sensor data from the back and arms may not provide sufficient distinct information to differentiate these activities effectively. Consequently, even with the fusion of lateral, interconnected, and analogous perspectives, the model struggles to capture the subtle variations between these tasks, revealing a drawback: the difficulty in identifying nuanced activity differences.

The comparison between the DSADS and OPPT datasets underscores that the Information Fusion GNN-ADG model is more effective for DSADS, where information fusion proves advantageous in scenarios involving complex, full-body movements. In DSADS, activities that require intricate coordination across multiple body parts—legs, arms, and torso—demanding a holistic integration of biomechanical relationships. The model's fusion of Interconnected, Analogous, and Lateral anatomical units excels here, capturing diverse perspectives like leg coordination, limb symmetry, and unilateral strength, enhanced by a cyclic training strategy that prevents overfitting to a single view. Conversely, OPPT focuses on upper-body tasks, which are more localized and often lack the full-body dynamics that maximize the fusion approach’s benefits. While OPPT activities leverage multifaceted upper-body interactions, many tasks are highly similar or subtle, and the limited sensor placement on the back and arms struggles to provide distinct data, reducing the model's ability to differentiate them. Thus, the Information Fusion GNN-ADG model is good at DSADS’s dynamic, full-body scenarios, where its comprehensive fusion mechanism enhances recognition of intricate movement patterns, whereas OPPT’s narrower scope limits its effectiveness.

\subsection{Ablation Study Analysis}

The GNN-ADG method introduces two loss functions, each addressing specific objectives during training. To evaluate the contributions of individual loss components in the proposed method, we perform a detailed correlation analysis between these losses and target user accuracy. Table~\ref{tab:correlation_results} presents the Pearson correlation coefficients and statistical significance levels between the loss components and target domain accuracy.

\begin{table}[htbp]
\centering
\caption{Correlation Analysis Between Loss Components and Target Accuracy}
\label{tab:correlation_results}
\begin{tabular}{lcc}
\toprule
\textbf{Loss Component} & \textbf{Pearson's r} & \textbf{p-value} \\
\midrule
Activity Classification Loss & $-0.946$ & $2.446 \times 10^{-74}$ \\
Source User Discrimination Loss & $-0.547$ & $4.682 \times 10^{-13}$ \\
\bottomrule
\end{tabular}
\end{table}

The analysis reveals two key insights:

\begin{itemize}
    \item \textbf{Activity Classification Loss} demonstrates a remarkably strong negative correlation with target accuracy. This near-perfect inverse relationship indicates that reductions in activity classification loss directly correspond to accuracy improvements, confirming its role as the primary driver of model performance. The statistical significance suggests this relationship is extremely unlikely to occur by chance.

    \item \textbf{Source User Discrimination Loss} shows a moderate but statistically significant negative correlation. While less pronounced than the activity loss relationship, this demonstrates that domain adaptation contributes meaningfully to final model performance through regularization effects. The persistent significance confirms the domain generalization mechanism effectively complements the primary classification task.
\end{itemize}

These results validate our dual-loss design strategy: the strong activity loss correlation confirms the GNN's ability to learn discriminative activity patterns, while the significant domain loss correlation demonstrates successful mitigation of domain shift. The inverse relationships for both components suggest simultaneous optimization of these losses creates synergistic effects -- reducing activity loss improves fundamental classification capability while minimizing domain loss enhances cross-domain generalizability.

\section{Conclusion}
\label{sec:conclusion}

In this paper, we introduced \textbf{Information Fusion GNN-ADG}, a novel Sensor Position-Aware Graph Neural Network with Adversarial Domain Generalization, to address cross-user variability in HAR. By combining anatomical correlation knowledge with adversarial learning, Information Fusion GNN-ADG effectively models spatial relationships among sensors placed on different anatomical parts, enabling robust generalization to unseen users. Our approach defines three anatomical units—\textbf{Interconnected}, \textbf{Analogous}, and \textbf{Lateral}—to capture universal patterns of coordinated movement. Rooted in biomechanical principles, these units reflect consistent sensor relationships across individuals, integrated through a unified graph with a cyclic training strategy. Adversarial learning, powered by a GRL, further extracts user-invariant features, reducing the impact of user-specific variations.

The experimental results on the OPPORTUNITY and DSADS datasets demonstrate that Information Fusion GNN-ADG outperforms state-of-the-art methods in cross-user activity recognition tasks. This highlights the effectiveness of our approach in addressing the challenges posed by cross-user variability, such as differences in body dynamics and individual behaviors. Importantly, Information Fusion GNN-ADG does not require target user data during training, making it highly practical for real-world applications where acquiring labeled data from new users is often infeasible.

Our work bridges the gap between biomechanical insights and computational models, offering a robust solution to the problem of cross-user variability in HAR systems. By fusing domain knowledge with machine learning techniques, we have shown that anatomical correlation knowledge can improve the generalization capability of HAR models. To the best of our knowledge, this is the first study to explore the shared and common knowledge of anatomical correlations for domain generalization in HAR through the lens of information fusion.

While Information Fusion GNN-ADG has demonstrated promising results, there are several directions for future research. First, extending the framework to incorporate multi-modal data, such as combining wearable sensor data with vision-based data, could further enhance the robustness and accuracy of activity recognition. Second, exploring the integration of temporal dynamics into the graph structure could provide a more comprehensive representation of human activities. Finally, investigating the application of Information Fusion GNN-ADG in other domains, such as healthcare monitoring, could validate its generalizability and utility in diverse real-world scenarios.

\bibliographystyle{elsarticle-num}
\bibliography{ref}

\begin{thebibliography}{10}
\expandafter\ifx\csname url\endcsname\relax
  \def\url#1{\texttt{#1}}\fi
\expandafter\ifx\csname urlprefix\endcsname\relax\def\urlprefix{URL }\fi
\expandafter\ifx\csname href\endcsname\relax
  \def\href#1#2{#2} \def\path#1{#1}\fi

\bibitem{bulling2014tutorial}
A.~Bulling, U.~Blanke, B.~Schiele, A tutorial on human activity recognition using body-worn inertial sensors, ACM Computing Surveys (CSUR) 46~(3) (2014) 1--33.

\bibitem{kwapisz2011activity}
J.~R. Kwapisz, G.~M. Weiss, S.~A. Moore, Activity recognition using cell phone accelerometers, ACM SigKDD Explorations Newsletter 12~(2) (2011) 74--82.

\bibitem{s24247975}
X.~Ye, K.~Sakurai, N.-K.~C. Nair, K.~I.-K. Wang, Machine learning techniques for sensor-based human activity recognition with data heterogeneity—a review, Sensors 24~(24) (2024).
\newblock \href {https://doi.org/10.3390/s24247975} {\path{doi:10.3390/s24247975}}.

\bibitem{ye2023cross}
X.~Ye, K.~I.-K. Wang, Cross-user activity recognition via temporal relation optimal transport, in: International Conference on Mobile and Ubiquitous Systems: Computing, Networking, and Services, Springer, 2023, pp. 355--374.

\bibitem{ordonez2016deep}
F.~J. Ord{\'o}{\~n}ez, D.~Roggen, Deep convolutional and lstm recurrent neural networks for multimodal wearable activity recognition, Sensors 16~(1) (2016) 115.

\bibitem{hammerla2016deep}
N.~Y. Hammerla, S.~Halloran, T.~Pl{\"o}tz, Deep, convolutional, and recurrent models for human activity recognition using wearables, arXiv preprint arXiv:1604.08880 (2016).

\bibitem{wang2018deep}
J.~Wang, V.~W. Zheng, Y.~Chen, M.~Huang, Deep transfer learning for cross-domain activity recognition, in: proceedings of the 3rd International Conference on Crowd Science and Engineering, 2018, pp. 1--8.

\bibitem{li2018learning}
D.~Li, Y.~Yang, Y.-Z. Song, T.~M. Hospedales, Learning to generalize: Meta-learning for domain generalization, in: Proceedings of the AAAI Conference on Artificial Intelligence, Vol.~32, 2018.

\bibitem{ahmad2021graph}
T.~Ahmad, L.~Jin, X.~Zhang, S.~Lai, G.~Tang, L.~Lin, Graph convolutional neural network for human action recognition: A comprehensive survey, IEEE Transactions on Artificial Intelligence 2~(2) (2021) 128--145.

\bibitem{ganin2015unsupervised}
Y.~Ganin, V.~Lempitsky, Unsupervised domain adaptation by backpropagation, in: International conference on machine learning, PMLR, 2015, pp. 1180--1189.

\bibitem{gil2023reducing}
M.~Gil-Mart{\'\i}n, J.~L{\'o}pez-Iniesta, F.~Fern{\'a}ndez-Mart{\'\i}nez, R.~San-Segundo, Reducing the impact of sensor orientation variability in human activity recognition using a consistent reference system, Sensors 23~(13) (2023) 5845.

\bibitem{soleimani2021cross}
E.~Soleimani, E.~Nazerfard, Cross-subject transfer learning in human activity recognition systems using generative adversarial networks, Neurocomputing 426 (2021) 26--34.

\bibitem{wang2024optimization}
S.~Wang, J.~Wang, H.~Xi, B.~Zhang, L.~Zhang, H.~Wei, Optimization-free test-time adaptation for cross-person activity recognition, Proceedings of the ACM on Interactive, Mobile, Wearable and Ubiquitous Technologies 7~(4) (2024) 1--27.

\bibitem{gupta2022human}
N.~Gupta, S.~K. Gupta, R.~K. Pathak, V.~Jain, P.~Rashidi, J.~S. Suri, Human activity recognition in artificial intelligence framework: a narrative review, Artificial intelligence review 55~(6) (2022) 4755--4808.

\bibitem{guo2021efficacy}
J.~Guo, U.~Kurup, M.~Shah, Efficacy of model fine-tuning for personalized dynamic gesture recognition, in: International Workshop on Deep Learning for Human Activity Recognition, Springer, 2021, pp. 99--110.

\bibitem{ye2025cross}
X.~Ye, K.~I.-K. Wang, Cross-user activity recognition using deep domain adaptation with temporal dependency information, IEEE Transactions on Instrumentation and Measurement 74 (2025) 1--15.

\bibitem{bento2023exploring}
N.~Bento, J.~Rebelo, A.~V. Carreiro, F.~Ravache, M.~Barandas, Exploring regularization methods for domain generalization in accelerometer-based human activity recognition, Sensors 23~(14) (2023) 6511.

\bibitem{thukral2025cross}
M.~Thukral, H.~Haresamudram, T.~Ploetz, Cross-domain har: Few-shot transfer learning for human activity recognition, ACM Transactions on Intelligent Systems and Technology 16~(1) (2025) 1--35.

\bibitem{ye2024deep}
X.~Ye, K.~I.-K. Wang, Deep generative domain adaptation with temporal relation attention mechanism for cross-user activity recognition, Pattern Recognition (2024) 110811.

\bibitem{gopalakrishnan2024comparative}
T.~Gopalakrishnan, N.~Wason, R.~J. Krishna, N.~Krishnaraj, Comparative analysis of fine-tuning i3d and slowfast networks for action recognition in surveillance videos, Engineering Proceedings 59~(1) (2024) 203.

\bibitem{genc2024human}
E.~Genc, M.~E. Yildirim, B.~S. Yucel, Human activity recognition with fine-tuned cnn-lstm, Journal of Electrical Engineering 75~(1) (2024) 8--13.

\bibitem{leite2022resource}
C.~F.~S. Leite, Y.~Xiao, Resource-efficient continual learning for sensor-based human activity recognition, ACM Transactions on Embedded Computing Systems 21~(6) (2022) 1--25.

\bibitem{YeSensor2024}
X.~Ye, K.~Sakurai, N.~C. Nair, K.~I. Wang, Machine learning techniques for sensor-based human activity recognition with data heterogeneity-a review, Sensors 24~(24) (2024) 7975.
\newblock \href {https://doi.org/10.3390/s24247975} {\path{doi:10.3390/s24247975}}.

\bibitem{farahani2021brief}
A.~Farahani, S.~Voghoei, K.~Rasheed, H.~R. Arabnia, A brief review of domain adaptation, Advances in data science and information engineering: proceedings from ICDATA 2020 and IKE 2020 (2021) 877--894.

\bibitem{xiaozhou2024genDATR}
X.~Ye, K.~I.-K. Wang, Deep generative domain adaptation with temporal relation knowledge for cross-user activity recognition, in: International Conference on Mobile and Ubiquitous Systems: Computing, Networking, and Services, 2024.

\bibitem{chakma2021activity}
A.~Chakma, A.~Z.~M. Faridee, M.~A. A.~H. Khan, N.~Roy, Activity recognition in wearables using adversarial multi-source domain adaptation, Smart Health 19 (2021) 100174.

\bibitem{hu2023swl}
R.~Hu, L.~Chen, S.~Miao, X.~Tang, Swl-adapt: An unsupervised domain adaptation model with sample weight learning for cross-user wearable human activity recognition, in: Proceedings of the AAAI Conference on artificial intelligence, Vol.~37, 2023, pp. 6012--6020.

\bibitem{napoli2024benchmark}
O.~Napoli, D.~Duarte, P.~Alves, D.~H.~P. Soto, H.~E. de~Oliveira, A.~Rocha, L.~Boccato, E.~Borin, A benchmark for domain adaptation and generalization in smartphone-based human activity recognition, Scientific Data 11~(1) (2024) 1192.

\bibitem{hong2024crosshar}
Z.~Hong, Z.~Li, S.~Zhong, W.~Lyu, H.~Wang, Y.~Ding, T.~He, D.~Zhang, Crosshar: Generalizing cross-dataset human activity recognition via hierarchical self-supervised pretraining, Proceedings of the ACM on Interactive, Mobile, Wearable and Ubiquitous Technologies 8~(2) (2024) 1--26.

\bibitem{li2021simple}
P.~Li, D.~Li, W.~Li, S.~Gong, Y.~Fu, T.~M. Hospedales, A simple feature augmentation for domain generalization, in: Proceedings of the IEEE/CVF international conference on computer vision, 2021, pp. 8886--8895.

\bibitem{volpi2018generalizing}
R.~Volpi, H.~Namkoong, O.~Sener, J.~Duchi, V.~Murino, S.~Savarese, Generalizing to unseen domains via adversarial data augmentation, in: Advances in Neural Information Processing Systems, Vol.~31, 2018.

\bibitem{lu2022domaininvariant}
W.~Lu, J.~Wang, H.~Li, Y.~Chen, X.~Xie, Domain-invariant feature exploration for domain generalization, Transactions on Machine Learning Research (2022).

\bibitem{parascandolo2021learning}
G.~Parascandolo, A.~Neitz, A.~Orvieto, L.~Gresele, B.~Sch{\"o}lkopf, Learning explanations that are hard to vary, in: International Conference on Learning Representations (ICLR 2021), OpenReview, 2021.

\bibitem{du2021adarnn}
Y.~Du, J.~Wang, W.~Feng, S.~Pan, T.~Qin, R.~Xu, C.~Wang, Adarnn: Adaptive learning and forecasting of time series, in: Proceedings of the 30th ACM international conference on information \& knowledge management, 2021, pp. 402--411.

\bibitem{huang2020self}
Z.~Huang, H.~Wang, E.~P. Xing, D.~Huang, Self-challenging improves cross-domain generalization, in: Computer vision--ECCV 2020: 16th European conference, Glasgow, UK, August 23--28, 2020, proceedings, part II 16, Springer, 2020, pp. 124--140.

\bibitem{wu2020comprehensive}
Z.~Wu, S.~Pan, F.~Chen, G.~Long, C.~Zhang, S.~Y. Philip, A comprehensive survey on graph neural networks, IEEE transactions on neural networks and learning systems 32~(1) (2020) 4--24.

\bibitem{seo2018structured}
Y.~Seo, M.~Defferrard, P.~Vandergheynst, X.~Bresson, Structured sequence modeling with graph convolutional recurrent networks, in: International Conference on Neural Information Processing, Springer, 2018, pp. 362--373.

\bibitem{si2019attention}
C.~Si, W.~Chen, W.~Wang, L.~Wang, T.~Tan, An attention enhanced graph convolutional lstm network for skeleton-based action recognition, in: Proceedings of the IEEE/CVF Conference on Computer Vision and Pattern Recognition, 2019, pp. 1227--1236.

\bibitem{shi2019two}
L.~Shi, Y.~Zhang, J.~Cheng, H.~Lu, Two-stream adaptive graph convolutional networks for skeleton-based action recognition, in: Proceedings of the IEEE/CVF Conference on Computer Vision and Pattern Recognition, 2019, pp. 12026--12035.

\bibitem{wieland2023tinygraphhar}
C.~Wieland, V.~Pankratius, Tinygraphhar: Enhancing human activity recognition with graph neural networks, in: 2023 IEEE World AI IoT Congress (AIIoT), IEEE, 2023, pp. 0047--0054.

\bibitem{ganin2016domain}
Y.~Ganin, E.~Ustinova, H.~Ajakan, P.~Germain, H.~Larochelle, F.~Laviolette, M.~March, V.~Lempitsky, Domain-adversarial training of neural networks, Journal of machine learning research 17~(59) (2016) 1--35.

\bibitem{li2018adaptive}
Y.~Li, N.~Wang, J.~Shi, X.~Hou, J.~Liu, Adaptive batch normalization for practical domain adaptation, Pattern Recognition 80 (2018) 109--117.

\bibitem{yao2017deepsense}
S.~Yao, S.~Hu, Y.~Zhao, A.~Zhang, T.~Abdelzaher, Deepsense: A unified deep learning framework for time-series mobile sensing data processing, in: Proceedings of the 26th International Conference on World Wide Web, 2017, pp. 351--360.

\bibitem{Kipf:2016tc}
T.~N. Kipf, M.~Welling, {Semi-Supervised Classification with Graph Convolutional Networks}, in: Proceedings of the 5th International Conference on Learning Representations, 2017.

\bibitem{chavarriaga2013opportunity}
R.~Chavarriaga, H.~Sagha, A.~Calatroni, S.~T. Digumarti, G.~Tr{\"o}ster, J.~d.~R. Mill{\'a}n, D.~Roggen, The opportunity challenge: A benchmark database for on-body sensor-based activity recognition, Pattern Recognition Letters 34~(15) (2013) 2033--2042.

\bibitem{barshan2014recognizing}
B.~Barshan, M.~C. Y{\"u}ksek, Recognizing daily and sports activities in two open source machine learning environments using body-worn sensor units, The Computer Journal 57~(11) (2014) 1649--1667.

\bibitem{zhang2018mixup}
H.~Zhang, M.~Cisse, Y.~N. Dauphin, D.~Lopez-Paz, mixup: Beyond empirical risk minimization, in: International Conference on Learning Representations, 2018.

\bibitem{lu2022domain}
W.~Lu, J.~Wang, H.~Li, Y.~Chen, X.~Xie, Domain-invariant feature exploration for domain generalization, arXiv preprint arXiv:2207.12020 (2022).

\end{thebibliography}

\end{document}